\documentclass[10pt,twocolumn,letterpaper]{article}
\usepackage{romannum}
\usepackage{ijcb}
\usepackage{times}
\usepackage{epsfig}
\usepackage{graphicx}
\usepackage{amsmath}
\usepackage{amssymb}
\usepackage{flexisym}
\usepackage{verbatim}
\usepackage{multicol}
\usepackage{subfigure}
\usepackage{cuted}
\usepackage{capt-of}
\usepackage{mathtools}
\newcommand{\Ham}{\mathrm{Ham}}
\newcommand{\R}{\mathbb{R}}
\newcommand\numberthis{\addtocounter{equation}{1}\tag{\theequation}}
\usepackage[linesnumbered,ruled,vlined]{algorithm2e}
\usepackage[hyphens,spaces,obeyspaces]{url}
\usepackage{multirow}
\usepackage{float}
\usepackage{caption}
\ifijcbfinal\fi
\ijcbfinalcopy

\title{\color{black}IHashNet: Iris Hashing Network 
based on 
efficient multi-index hashing}

\author{Avantika Singh\\
IIT Mandi\\
{\tt\small aavantikatomar@gmail.com}
\and
Chirag Vashist\\
IIT Mandi\\
{\tt\small chiragvashist007@gmail.com}
\and
Pratyush Gaurav\\
IIT Mandi\\
{\tt\small pratyushgauravgo@gmail.com}
\and
Aditya Nigam\\
IIT Mandi\\
{\tt\small aditya@iitmandi.ac.in}
\and
Rameshwar Pratap\\
IIT Mandi\\
{\tt\small rameshwar@iitmandi.ac.in}

}

\begin{document}

\maketitle

\begin{strip}\centering
\includegraphics[width=0.9\textwidth,height=0.288\textwidth]{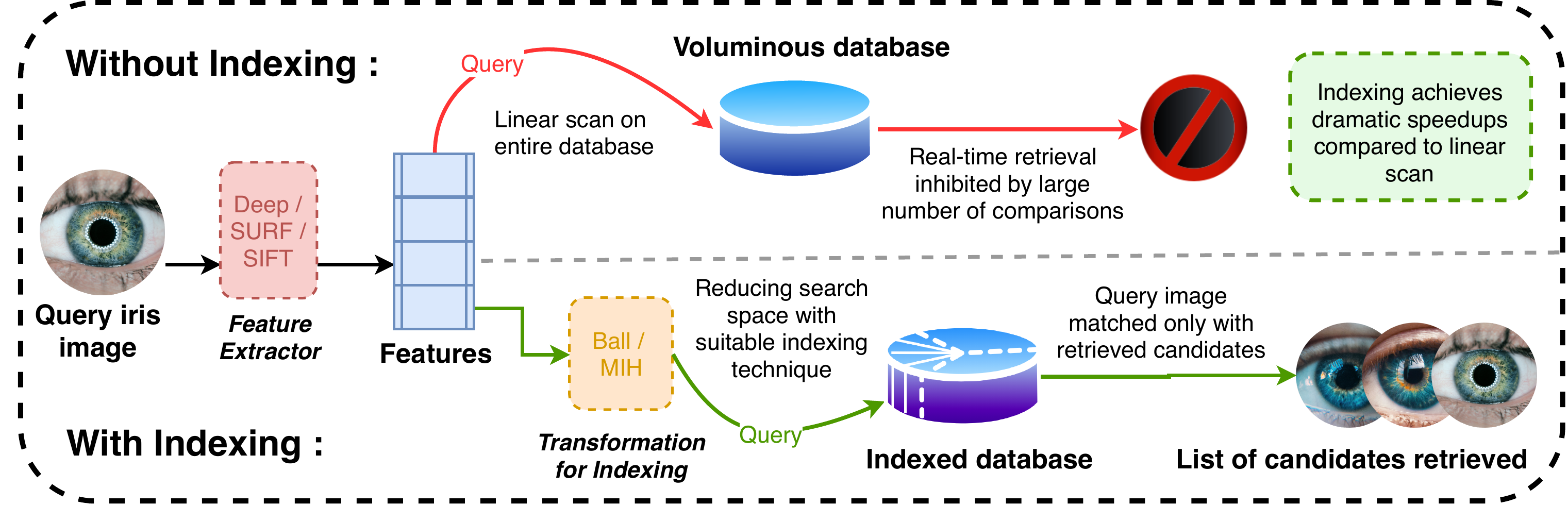}
		
\captionof{figure}{ Depicting the advantage of indexing Iris image. The upper part depicts the scenario in which the extracted iris features are directly stored in the database without indexing. Mostly query  triggers a linear scan on the entire database {\color{black}(except few works like \cite{bowyer1})} which is impractical in real life scenarios owing to high identification time.
 The second scenario employs indexing to store iris images. 
 \label{iriss are _anatomy}}
\end{strip}	
\begin{abstract}
   Massive biometric deployments are pervasive in today's world. But despite the high accuracy of biometric systems, their computational efficiency degrades drastically with an increase in the database size. Thus, it is essential to index them. An ideal indexing scheme needs to generate codes that preserve the intra-subject similarity as well as inter-subject dissimilarity. 
   Here, in this paper, we propose an iris indexing scheme using real-valued deep iris features binarized to iris bar codes (IBC) compatible with the indexing structure. Firstly, for extracting robust iris features, we have designed a network utilizing the domain knowledge of ordinal filtering and learning their nonlinear combinations. Later these real-valued features are binarized. Finally, for indexing the iris dataset, we have proposed $M_{com}$ loss that can transform the binary feature into an 
   improved 
   feature compatible with Multi-Index Hashing scheme. This $M_{com}$ loss function ensures the hamming distance equally distributed among all the contiguous disjoint sub-strings. To the best of our knowledge, this is the first work in the iris indexing domain that presents an end-to-end iris indexing structure. Experimental results on four datasets are presented to depict the efficacy of the proposed approach.
   

\end{abstract}

\section{Introduction} Due to the digitization and expeditious deployment of biometric systems like Aadhar(India)~\cite{aadhar} and MyKad(Malaysia)~\cite{mykad},  it is very crucial to facilitate rapid search.
\color{black}
Currently around $1.2$ billion people in India are enrolled under the Aadhar scheme. Mumbai is the most populous city in India with inhabitants around $18$ million, out of which around $9.5$ million population is males and around $8.5$ million population is females \footnote{\url{https://www.census2011.co.in/census/city/365-mumbai.html}}. One can observe that even after the entire country's population got divided on the basis of their location and gender, the resulting population will be very large (may be of the order of millions). Also, although Aadhar system is handling tens of millions of authentications on daily basis, utilizing huge amount of computing resources and data level parallelism.  It has been reported that Aadhar does 1:1 matching (i.e. verification) and never performs 1:N matching (i.e. recognition)\footnote{\url{ https://uidai.gov.in/aadhaar-eco-system/authentication-ecosystem.html}}. It restricts recognition (1:N), except while checking duplicates while issuing new Aadhar card which take huge amount of time (may be in days). While the proposed indexing system can facilitate recognition (1:N), and ensures fast and accurate retrieval. In general Aadhar operations are concealed from general public. The proposed indexing strategy could be used as a sub categorical  parameter for improving the indexing under the main categorical distinguishing parameters like gender and location. This system can facilitate recognition (1:N) and ensures fast and accurate retrieval.
\color{black}
Out of several biometric traits, the iris is regarded as one of the best biometric modality owing to its high discriminative power due to highly textural patterns \cite{iris_nor} and less susceptibility to various spoofing attacks.

\textbf{Problem Statement: } The goal of any indexing technique is perform authentication with minimum comparisons. It can be achieved by only considering the semantically neighboring query indices instead of the entire database. Figure.\ref{iriss are _anatomy} presents a general overview of iris identification under two different scenarios. 

\textbf{Challenges: } The fuzziness of the iris dataset, which means the gallery and the probe iris image belonging to the same subject, appears similar but never identical, makes the indexing technique even more challenging. Another critical issue is the trade-off between accuracy and efficiency, which leads to performance deterioration when fast searching is performed. 

\begin{figure*}[!htp]
	\begin{center}
		\includegraphics[width=0.85\linewidth]{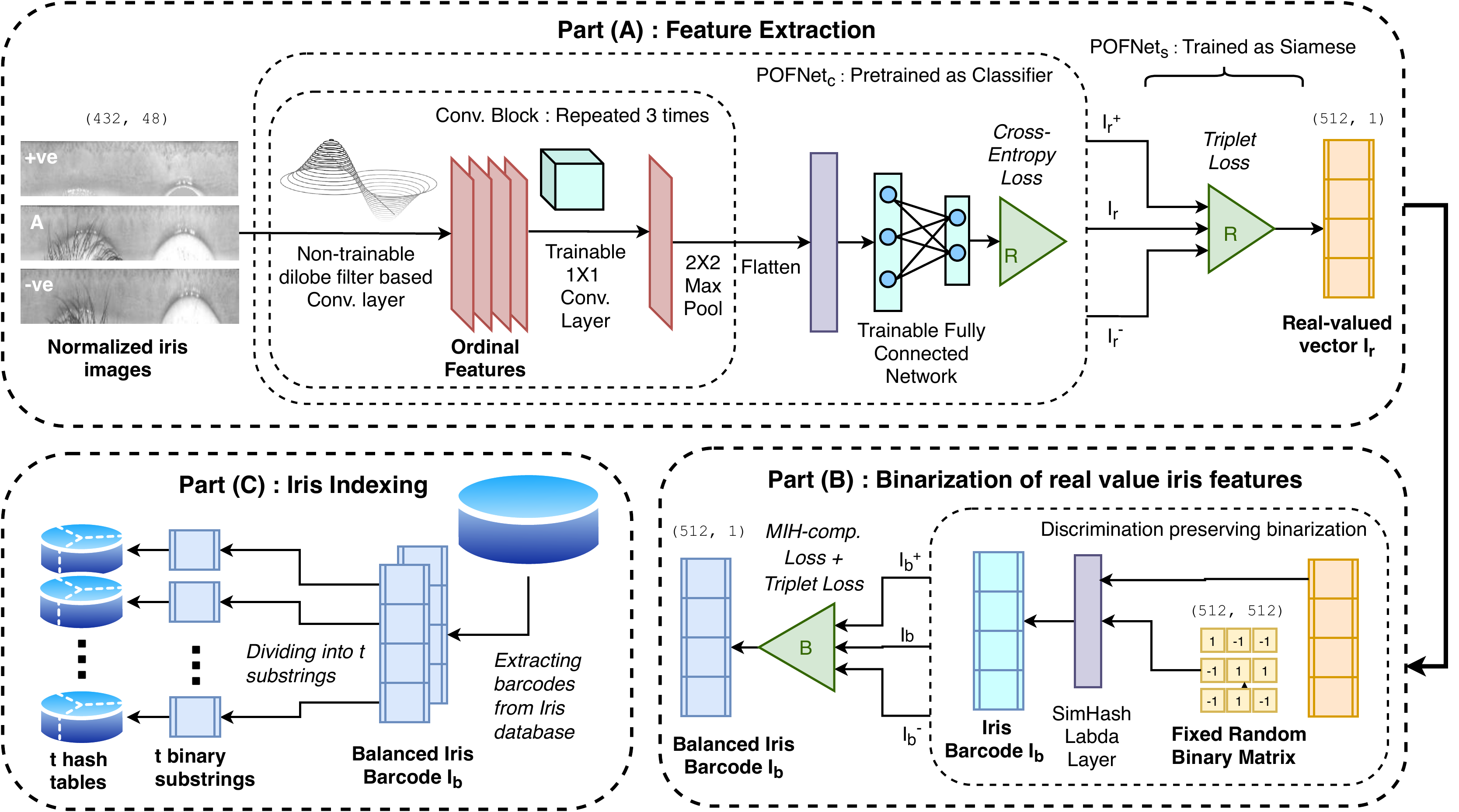}
	\end{center}
	\caption{Proposed Methodology. Part (A) depicts the network employed to extract real-valued features from normalized iris images, using patch-level ordinal filters. Part (B) depicts binarization of real-valued features obtained from Part (A) using SimHash, while preserving their discriminative features. The barcodes subsequently obtained are bit-balanced by training the network on MIH compatible loss ($M_{com}$ loss) on the barcodes along with the Triplet loss on the real-valued features. Part (C) describes Multi-index strategy. Using the network trained in Part(A) and Part(B), iris images in the database are converted to their corresponding bit-balanced barcodes. Each barcode is divided into $t$ contiguous and disjoint substrings, with each one of the substrings populating its corresponding hash tables.}
	\label{iris_method}
\end{figure*}


\textbf{Related work: } Several works have been done in the literature for iris recognition \cite{ajay1,ajay2} but the work done in the field of iris indexing is limited. 
Daugman ~\cite{daugman} does one of the initial work in this field proposing a search algorithm on iris codes based on a beacon guided search using multiple colliding segment principle but requires complex memory management.  In another notable work, Jayaraman et al. ~\cite{umarani} developed a method for indexing large scale iris database based on iris color. However, the proposed technique ~\cite{umarani} is not suitable owing to its high dimensionality.  In one of the work, Mukherjee and Ross ~\cite{ross} proposed an indexing strategy based on SPLDH (signed pixel-level difference histogram). However, the major challenge in this method ~\cite{ross} is to determine the histogram dimensionality. Recently, Rathgeb et al. ~\cite{rathgeb} study the concept of a bloom filter for indexing iris dataset giving promising results. To the best of our knowledge, the work done by Ahmed et al. \cite{latest_iris} is the current state-of-the-art in iris indexing. In their work, they have proposed an indexing strategy based on hyper-planes.

\textbf{Contribution: } 
 The key contributions  are as follows:\\
{\color{black}
\textbf {\textit{1) Image to Real value feature extraction}}: A novel  deep neural architecture (POFNet) based on ordinal  measures for extracting patch level discriminative iris information is proposed as discussed in more detail under heading patch-level ordinal features (POFNet) in section \ref{methodology}.\\
\textbf {\textit{2) Optimized MIH }}: An optimized version of the Multi index hashing (MIH)  scheme  is proposed as defined in Algorithm 1. Table  \ref{tab_hit} further indicates the superiority of optimized MIH  in terms of penetration rate and retrieval time in comparison to naive MIH implementation. \\ 
\textbf{ \textit{3) $M_{com}$ loss}:}
A novel $M_{com}$ loss has been proposed that transformed binary iris features into an improved feature compatible with Multi-Index Hashing scheme. This loss function ensures the hamming distance equally distributed among all the contiguous disjoint sub-strings.}

\section{Proposed methodology} \label{methodology}

 Highly distributed iris structures make it one of the best biometric modality. However, detecting these microtextural iris patterns is a challenging issue, especially due to the presence of obstructions like eyelids, eyelashes, and all. Traditionally several feature descriptors like LBP (Local Binary Pattern),  LoG (Laplacian of gaussian), DWT (Discrete Wavelet Transform), Gabor filter bank, and relational measures were used for extracting robust iris features. Recently, it has been proved that CNN's(Convolutional neural networks) 
are a good choice for learning robust feature descriptors. CNN's are used in iris domain in a variety of tasks like iris recognition, iris segmentation, iris spoof detection. This inspired us to extract robust iris features using CNN's. 
\subsection{Iris deep feature extraction network} Due to the involvement of a large number of training parameters, often training a deep network is a cumbersome job. Very recently, LBP \cite{vishnu} based CNN network is proposed for face recognition in which most of the filters are fixed to reduce the computational complexity of the deep network. This inspired us to use an amalgam of traditional iris domain knowledge along with deep learning. Thus, for learning robust iris features, we have constructed a deep network based on patch level ordinal features (POFNet). It is well established in literature ~\cite{iris_ordinal} that ordinal measures among neighboring image pixels exhibit some sort of stability under varying illumination conditions.  Ordinal measures encode qualitative information rather than its quantitative values. Thus, in the designing of POFNet, most of the filters are made non-trainable (fixed) which are based on patch level ordinal filters, and to learn the best possible combination of these non-trainable filter outputs, 1*1 trainable conv is used. Once the robust real-valued iris features are obtained, the next step is to binarized them for effective hashing. For binarization, we have proposed a discrimination preserving binarization (DPB) layer.
 Figure.\ref{iris_method} provides a general overview of the proposed methodology, while the below subsections explain it in more detail.
 
\textbf{(A) Patch-level ordinal features(POFNet):} Due to the various challenging issues suffered by the iris recognition system, matching is mostly performed on the normalized iris-image rather than on original iris-image. Thus, in our work,  we have used normalized iris images, which are segmented first by using the algorithm proposed by ~\cite{iris_seg} and later normalized by using an algorithm proposed by ~\cite{iris_nor}. For learning robust iris features, we have constructed a deep network based on patch level ordinal features (POFNet). The advantage of dilobe ordinal filters (DOF) in iris recognition has been well established by \cite{iris_ordinal}. DOF has a positive and negative Gaussian lobe. It is specifically designed in terms of orientation, scale, location, and distance to measure the ordinal relationship between iris images.  It works in a patch-wise manner, comparing one patch with other and thus extracting \textbf{domain specific} highly textural iris features. Mathematically, a dilobe filter can be  represented as:
\color{black}
\begin{equation}
\begin{aligned}
C_p \ast \frac{1}{\sqrt{2\pi}\delta_{p}}e^{\frac{-(X-\mu_{p})^2}{2\delta_{p}^2}}{-} C_n \ast \frac{1}{\sqrt{2\pi}\delta_{n}}e^{\frac{-(X-\mu_{n})^2}{2\delta_{n}^2}}
\end{aligned}
\end{equation}
\color{black}
where $C_{p} \in (0\ to\ 1)$ and $C_{n}=C_{p}$, used to make zero-sum of the dilobe filter. The 2-D vector $X$ represents the spatial location with  $\mu_{p}, \mu_{n}$ and $\delta_{p}, \delta_{n}$ denote the mean and scale of positive and negative 2D-Gaussian filter. For more details  kindly refer \cite{iris_ordinal}.

\textbf{Network specifications:}
\begin{table}[!htp]
    \centering
    \color{black}
    \small{
    \caption{Network specifications with output shapes for IITD dataset. Output of each layer differs in dimension for every dataset owing to the different sizes of normalized iris images.}
    \label{tab_real}
    \small{}
    \begin{tabular}{|p{5.0cm}|p{2.0cm}|}
        \hline
        \textbf{Layer Name} & \textbf{Output Shape} \\ \hline
        Input Layer & (48, 432, 1) \\ \hline
        $3$X$5$ Dilobe Conv. | ReLU Activation & (48, 432, 8) \\ \hline
        $1$X$1$ Conv. Layer | ReLU Activation & (48, 432, 1) \\ \hline
        Batch Normalization & (48, 432, 1) \\ \hline
        $3$X$5$ Dilobe Conv. | ReLU Activation & (48, 432, 16) \\ \hline
        $1$X$1$ Conv. | ReLU Activation  & (48, 432, 1) \\ \hline
        $2$X$2$ Max-pooling (stride=2) & (24, 216, 1) \\ \hline
        Batch Normalization & (24, 216, 1) \\ \hline
        $3$X$5$ Dilobe Conv. | ReLU Activation & (24, 216, 32) \\ \hline
        $1$X$1$ Conv. | ReLU Activation & (24, 216, 1) \\ \hline
        $2$X$2$ Max-pooling (stride=2) & (12, 108, 1) \\ \hline
        Batch Normalization & (12, 108, 1) \\ \hline
        Dense | ReLU Activation  & (256) \\ \hline
        Dense | ReLU Activation & (512) \\ \hline

    \end{tabular}}
\end{table}

The proposed POFNet consists of three non-trainable dilobe convolution layers(DCL), each composed of  fixed ($3 * 5$) filters of varying number, orientation and scale.
{\color{black}Max pooling is  mainly done to help  circumvent over-fitting by providing an abstracted form of the representation. In the first DCL we have not used max pooling because initially we want to retain all the values of the feature map so that subsequent layers can get maximum information and can extract highly discriminative iris features. Afterwards, we have used max-pooling to provide basic translational invariance and reduce the dimensions of the features.}

\textbf{Network justifications:} We have extracted patch-level ordinal features which are more focused on local areas and thus can better represent the local structure of iris. Furthermore, they help in extracting robust iris features ~\cite{iris_ordinal}. Moreover, the usage of $1*1$ Conv helps in learning the best non-linear combination of patch-level ordinal features. For deciding the network parameters, we have used the domain knowledge~\cite{iris_ordinal} and verified it experimentally.
\begin{figure*}[!htp]
    \begin{center}
        \includegraphics[width=0.9\linewidth,height=0.51\linewidth]{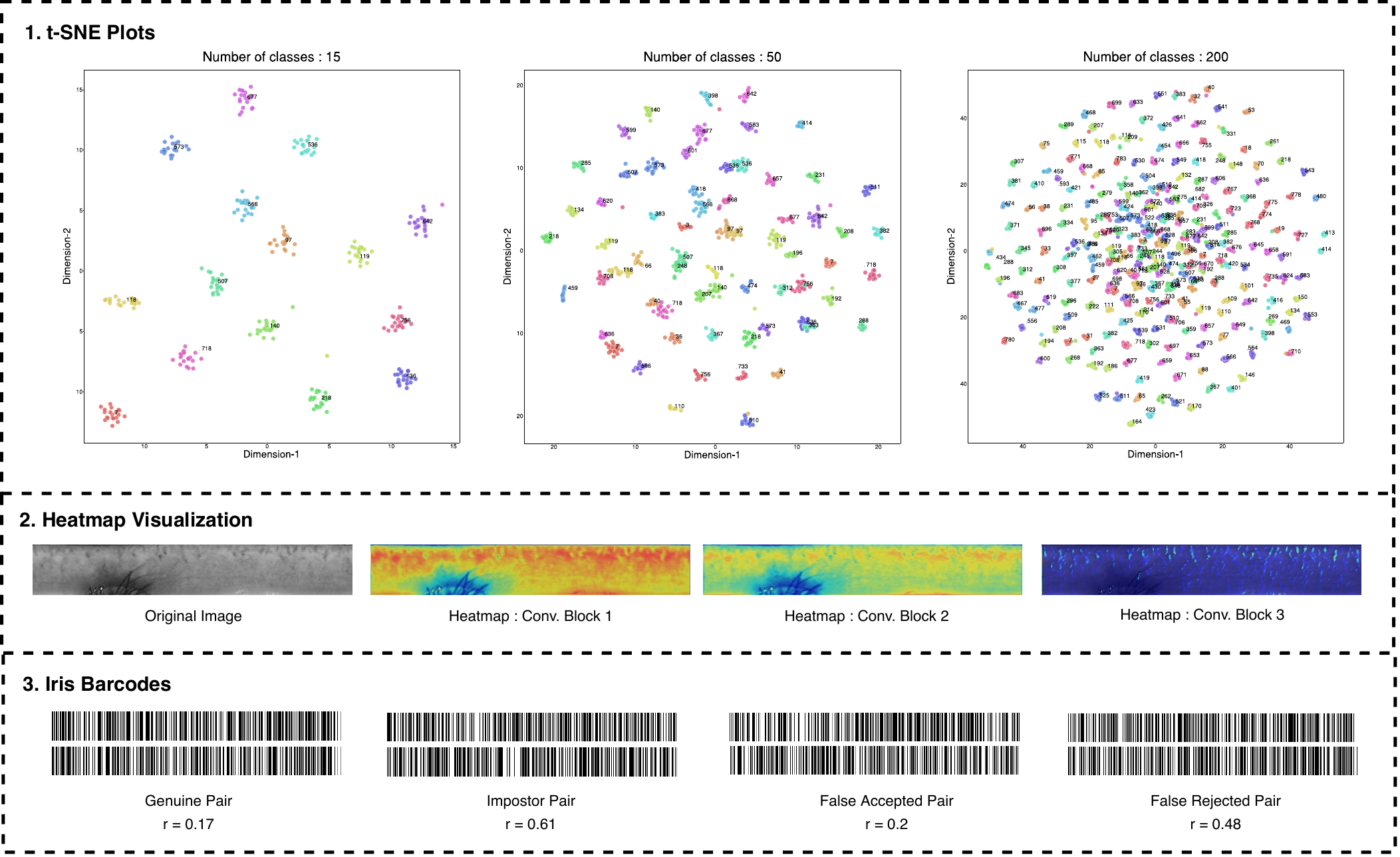}
    \end{center}
    \caption{Row-1 shows the t-SNE plot of various classes chosen randomly from CASIA-Lamp dataset. Row-2 depicts the heap-map generated from our deep network after various Convolutional Blocks (red and yellow indicate high attention areas). Row-3 shows the binarized vector pairs (iris barcodes) for genuine, imposter, false rejection and false acceptance (r is hamming distance) }
    \label{iris_features}
\end{figure*}

\textbf{(B) Network training strategy:} We have trained our network end-to-end  modularly. We have used  Python augmentor library \cite{python_aug} for data augmentation and used operations like random rotation, zooming, and random distortion. As shown in Part(A) Figure.\ref{iris_method}, we have first pre-trained our network as a classifier using cross-entropy loss and then applied triplet loss as defined in ~\cite{facenet} using a Siamese architecture.

\textbf{Triplet based training(TT):}
\color{black}{ Let the network 
$\mathcal{F}$ embeds an iris image $I_r$ as $\mathcal{F}(I_r)$. In triplet loss the goal is to ensure that the iris image $I_r^a$ (anchor) of a particular person is close to all iris images $I_r^+$
(positive) extracted from the same person but captured at different times and at the same time far apart from iris image  $I_r^-$(negative) extracted from  all the other persons. Thus, the objective is:
\begin{equation}
\left\lVert \mathcal{F}(I_r^a)-\mathcal{F}(I_r^+) \right\rVert_2^2 + \alpha <
\left\lVert \mathcal{F}(I_r^a)-\mathcal{F}(I_r^-) \right\rVert_2^2
\end{equation}
here $\alpha$ is margin. Let $N$
is the total number of possible triplets in the training set then the overall loss function is defined as:
\begin{equation}
L = \sum_i^N \Bigg[ \left\lVert \mathcal{F}(I_r^a)-\mathcal{F}(I_r^+) \right\rVert_2^2 -
\left\lVert \mathcal{F}(I_r^a)-\mathcal{F}(I_r^-) \right\rVert_2^2 + \alpha \Bigg]
\end{equation}
}
\color{black}
For effective network training we have used dynamic adaptive margin with hard negative mining as defined in ~\cite{daksh_18}.

\textbf{Hard negative mining:} random selection of triplets can easily ignore challenging one and thus can degrade the network performance. A triplet is said to be hard when the distance  $d_n$ (between negative, anchor embedding) and  $d_p$ (between positive, anchor embedding) is less than the margin ($\alpha$). To compute such embeddings is difficult. Thus, while making a batch, we randomly choose $1000$ triplets, and only those triplets are chosen, which satisfies the following condition:
\begin{equation}
d_n - d_p \leq \alpha
\end{equation}

\textbf{Dynamic adaptive margin:} as the training progress, the number of suitable triplets decreases due to a decrease in $d_p$  and an increase in $d_n$.  To circumvent this situation, we increase the margin($\alpha$) by 0.05 after the number of hard triplets become less than 10\% of the total mined triplets.

\textbf{Feature visualization:}  t-SNE plots and heat-maps are used for visualizing the network. It is clearly evident from Row1 Figure.\ref{iris_features} that triplet loss with adaptive margin is helping in clustering similar class samples together. Further, with the increase in the number of  classes, the distance between the distinct classes decreases as expected, and most of them fall in the center depicting a shape of a 2-D Gaussian. We have also analyzed the heat-maps generated through our network, as shown in  Row2 Figure.\ref{iris_features}. \color{black}{It is evident from Figure.\ref{iris_features} that our network is paying attention to the highly textural iris micro-structures which are highly encoded and are less interpretable.
}

\color{black}
\subsection{Discrimination preserving binarization}
 Voluminous datasets quickly strain memory and make disk access slow. Thus, it is highly desirable to develop search methods whose time and space complexity do not blow up with the input dimension and simultaneously designing a mapping strategy that maps high dimensional feature descriptors to a compact but discriminative representation. One possible solution is to use binary feature descriptors that not only fits quite well with the existing hamming ball based hashing strategies but also maximizes the information content of each bit. This inspired us to use binarized iris features for hashing purposes instead of using real-valued iris features extracted from our network. For that, we have designed a lambda layer named as discrimination preserving binarization layer (DPB Layer) performing signed random projection (SimHash~\cite{simhash}), as shown in Part(B) of Figure.\ref{iris_method}. 
 Using SimHash~\cite{simhash} we convert real valued iris features say $I_{re}\in \R^n$ into iris bar codes (binary feature) say $I_{b} \in \{0,1\}^k$  as follows:-
 
 We generates a random vector ${p}\in \{-1, +1\}^n$,  
where each component is generated from $\{-1, +1\}$ with equal probability (i.e  $0.5$), and stores only the sign of the projected data. That is,
\begin{equation}
 SimHash^{(p)}(I_{r})=\begin{cases}
    1, & \text{if $\langle I_{r}, {p}\rangle $} \geq 0.\\
    0, & \text{otherwise}.
  \end{cases}
\end{equation}
For a pair of vectors $I_{r}^{(1)}, I_{r}^{(2)} \in \R^{n}$,   
due to~\cite{GoemansW95} we have the   following guarantee  
\[
 \Pr[SimHash^{(p)}(I_{r}^{(1)}) = SimHash^{(p)}(I_{r}^{(2)})]=1-\frac{\theta}{\pi}, 
\]
where $\theta=\cos^{-1} \left( \langle I_{r}^{(1)}, I_{r}^{(2)} \rangle / ||I_{r}^{(1)}||_2\cdot{}||I_{r}^{(2)}||_2 \right)$. 
We repeat the above hashing experiments $k$ times and obtain their signature binary vectors 
$I_{b}^{(1)}, I_{b}^{(2)} \in \{0, 1\}^{k}$. 
The angular similarity between two real valued Iris features $ I_{r}^{(1)}, I_{r}^{(2)}$   can be computed using the
hamming distance between their respective Iris bar codes $I_{b}^{(1)}, I_{b}^{(2)}\in \{0, 1\}^{k}$ using the following equation (due to~\cite{simhash}).
\begin{equation}
 \cos(I_{r}^{(1)}, I_{r}^{(2)})=\cos\left[\left(\frac{\pi}{k} \right)  \Ham(I_{b}^{(1)}, I_{b}^{(2)}) \right]  \numberthis\label{eq:cosine_hamming}.
\end{equation}

    SimHash~\cite{simhash} preserves the angular distances between real-valued iris feature, which can be estimated from their respective iris bar codes due to Equation~\ref{eq:cosine_hamming}.  Discriminative preserving property is also depicted by iris barcodes of genuine pairs and imposter pairs shown in Row3 Figure.\ref{iris_features}. This is the reason for using it to convert real-valued feature vectors into binary forms instead of directly learning binary descriptors from the deep network. 

\subsection{Indexing structure for binarized iris features} Naive approach for claiming the identity of the query subject is to search the entire database. However, when the database is voluminous, the processing time grows linearly with its size. Thus, it is important to search for strategies that grow sublinear instead of linear in terms of time with database size. Here,  for indexing  iris bar codes we have used four strategies: (1) Ball-tree~\cite{ball_tree} (2) Multi-Index Hashing(MIH)~\cite{mih} (3) MIH optimized(4) $M_{com}$ MIH.
\newline
\textbf{ (A) Multi-Index hashing:} Most of the existing deep learning-based methods for hashing aims at improving the performance in terms of hit rate and penetration rate but gives less emphasis on the computational complexity which is essential for real-time retrieval. Multi index hashing (MIH) aims at improving the time complexity while maintaining the performance. 

\textbf{General:} It works as follows in our case:\\
\textbf{[1]} During enrollment for creating a database, multiple hash tables are built. Each hash table corresponds to the disjoint subset of iris barcode i.e. $I_{bar}$. Here the barcodes are indexed $t$ times into $t$ different hash tables as shown in Part(C) Figure.\ref{iris_method}.\\
     \textbf{[2]} During authentication given a query iris image say $I_q$,  first step is to obtain iris bar code corresponding to it, and further, it is divided into $t$ disjoint substring. \color{black}{In our experimentation, for $512$ bit iris bar code we have divided it into $16$ disjoint substrings. This setting is empirically set and experimentally verified}. \color{black} Then MIH searches each substring hash table that is within a hamming distance of $r/t$ to find $r$ neighbors of $I_q$ say $N_{rq}$.\\
     \textbf{[3]} Since the MIH considers only the substring not the complete barcode, all the elements in $N_{rq}$ are not true $r$ neighbors of $I_q$. The last step is to check for true $r$ neighbors by checking the complete binary string.\\
     \textbf{Optimized variant of Multi-Index hashing:}
To further improve the performance of MIH, we have proposed a variant of MIH, as described in Algorithm~\ref{algo_1}.
\newline
\begin{algorithm}[]
    \caption{Optimized Multi Index hashing}
    \label{algo_1}
    \KwIn{Binarized iris feature descriptor say $I_b$ corresponding to query input image and ball radius $r$}
    \KwOut{List of candidates that are true $r$ neighbours of $I_b$. }
    Divide input binary descriptor $I_b$ into $t$ disjoint substring. \\
    \For{$i \gets 1$ to $t$}{
        Prepare $candidates_i$: In $i^{th}$ hash table, query for $i^{th}$ substring of $I_b$ using radius $\left \lfloor{r/t}\right \rfloor$. \\
        Prune candidates in $candidates_i$ that are not true $r$ neighbour. \\
        \uIf{$answer \in $ $candidates_i$}{
            \bf{break};
        }
    }
\end{algorithm}
\textbf {(B) Indexing reconcilable iris bar code generation:} It has been proved that the MIH time complexity is sub-linear ~\cite{mih} when the binary codes are distributed uniformly. In the case of MIH r-neighbor candidates are detected first and then in the next step detected r-neighbor candidates are pruned to find the true r-neighbor candidates. In this scenario, the number of r-neighbor candidates is always greater than the true r neighbors candidates. Thus, the computational time complexity can be reduced if the number of r-neighbors is actually equivalent to true r-neighbors. It has been proven \cite{mih} that this can be done when the hamming distance of each corresponding sub-string pair is equal for any two binary strings, and is equal to $r/m$.
Where $r$ is the full hamming distance between the two vectors, and $m$ is the number of sub-strings.  \cite{deep_indexing}. We introduce a new loss function $M_{com}$:-

\begin{equation}
{M_{com}}(h_i,h_j)=\sqrt{\frac{1}{m}\sum_{m\textprime = 1}^{m} (cosine(h_i^{m\textprime},h_j^{m\textprime})-r\textprime)^2}
\end{equation}

where r\textprime = $r/m$, $h_i^m\textprime$ and $h_j^m\textprime$ represent the $m\textprime$ th substrings of hash vectors $h_i$ and $h_j$ and $cosine(a,b)$ represents the cosine distance between any two vectors $a$ and $b$.
For ease of optimization, we replace integer constraints in DPB layer with range constraints using tanh, which makes hash vectors real-valued, as suggested in~\cite{range}. For the purpose of training, we have used $M_{com}$ loss in addition with Triplet loss with $M_{com}$ loss having $10\%$ of the total weightage whereas, for indexing, we have used our optimized MIH approach to fetch candidates after the network is trained and named it as $M_{com}$ MIH. As depicted in point 3 of 
Figure.\ref{iris_features}) iris barcodes are generated by replicating 1D-iris binary code bits into the columns.

\subsection {Iris barcode-based retrieval} During the identification when an image is queried to the database, all the images that have close proximity (in terms of hamming distance in barcode space) with the queried image is retrieved from the database as follows:
\newline
[1] Given a query iris image
firstly extract its binarized features through our end to end network comprising of POFNet and discrimination preserving binarization (DBP) layer, which is trained using triplet loss and $M_{com}$ loss.
\newline
[2] Once binarized features are obtained, traverse the hash table built by using any indexing strategies such as Ball-tree, MIH, optimized MIH, or $M_{com}$ MIH to retrieve r-neighbors of the given query image.
\newline
[3] Compare the query image with retrieved r-neighbors.
\newline
[4] Return the probable matches.
    
\section{Experimental analysis} This section is focused on demonstrating the performance of the proposed methodology on different data sets considered.  The detailed explanation about the database specifications, experimental setup, along with result analysis, is presented in the following subsections.
\subsection{Database specification} We have tested our proposed methodology on four datasets (1) CASIA Lamp ~\cite{lamp} (2) CASIA-V3 Interval ~\cite{interval} (3) IITD-V1 Iris dataset ~\cite{iitd} (4) IITK Iris dataset. Out of all these four datasets, all datasets are publicly available except the IITK dataset. The specification of the 
Table \ref{tab_database} shows the detail description of the various dataset used in our experimentation work along with the number of queries and stored images corresponding to each dataset. \color{black} It should be noted that stored images act as training images while training the  
 deep architecture (POFNet).
\color{black}

 \begin{table}[]
     \centering
     \color{black}
     \small{
     \caption{Database used in the experiments}
     \label{tab_database}
     \begin{tabular}{|p{1.3cm}|p{1.3cm}|p{1.15cm}|p{1.0cm}|p{1.4cm}|}
         \hline
         & \textbf{IITK} & \textbf{ Interval} & \textbf{IITD} & \textbf{Lamp} \\ \hline
         \textbf{Total images} & $20420$ & $2555$ & $1120$ & $15660$ \\ \hline
         \textbf{No. of classes} & $2042$ & $349$ & $224$ & $819$ \\ \hline
         \textbf{Images per class} & $10$ & $2$ to $5$ & $5$ & $20$ \\ \hline
         \textbf{Stored images} & $8168$ & $1022$ & $450$ & $6264$ \\ \hline
         \textbf{Query images} & $12252$ & $1533$ & $670$ & $9396$ \\ \hline
         \textbf{Strip size} &$256\times64$ & $256\times64$ &$ 432 \times48$ &  $512\times 80$ \\ \hline
         \textbf{Genuine matching} & $49008$ & $4541$ & $1344$ & $78300$ \\ \hline
         \textbf{Imposter matching} & $100025328$ & $1458133$ & $299712$ & $61230600$ \\ \hline
     \end{tabular}}
 \end{table}
\subsection{Experimental setup}  POFNet feature extraction network is implemented in Python using Keras ~\cite{keras} library. For evaluating the proposed indexing mechanism and for training the network, a PC having Xenon (R) processor with $32$ GB RAM and $12$ GB on card RAM on NVIDIA Tesla K40C GPU has been used.

\textbf{Experimental protocol:} Since there is no standard training and testing protocol associated with the datasets considered. We have used the experimentation protocol as defined in the current state-of-the-art ~\cite{latest_iris}, in which $40\%$ of the dataset is used for training the network and rest for testing.
\color{black} (It should be noted that this dataset partitioning is not subject independent). \color{black} For analyzing the indexing performance, we have used three different bit vector sizes i.e., $128$, $256$, and $512$, respectively, but while reporting the performance, we have used only $512$ bit vectors (as it is performing best). For analyzing the performance of real-valued features extracted by the POFNet, performance comparison is done in terms of EER (Equal Error Rate), FAR (False Acceptance Rate), FRR (False Rejection Rate), DI (Decidability Index), Recall and Precision. For analyzing the performance of iris barcodes, we have used EER, FAR, FRR and DI. Further, for analyzing the performance of iris indexing, hit-rate and penetration rate have been used. For details on performance parameters refer ~\cite{latest_iris}.
\begin{table}[]
\centering
\small{
\caption{Performance with real value feature vectors(R) and with binarized iris barcodes(B) (FA and FR computed at Equal error rate (EER))}
\label{tab_real}
\begin{tabular}{|p{0.8cm}|p{0.45cm}|p{0.45cm}|p{0.45cm}|p{0.45cm}|p{0.45cm}|p{0.45cm}|p{0.45cm}|p{0.45cm}|}
\hline
\multirow{2}{*}{} & \multicolumn{2}{l|}{\textbf{Lamp}} & \multicolumn{2}{l|}{\textbf{Interval}} & \multicolumn{2}{l|}{\textbf{IITD}} & \multicolumn{2}{l|}{\textbf{IITK}} \\ \cline{2-9} 
 & \textbf{R} & \textbf{B} & \textbf{R} & \textbf{B} & \textbf{R} & \textbf{B} & \textbf{R} & \textbf{B} \\ \hline
\textbf{FAR\%} & 1.50 & 1.49 & 2.54 & 2.76 & 1.11 & 1.33 & 1.83 & 1.93 \\ \hline
\textbf{FRR\%} & 1.45 & 1.56 & 2.53 & 2.77 & 1.12 & 1.34 & 1.83 & 1.93 \\ \hline
\textbf{EER\%} & 1.48 & 1.54 & 2.53 & 2.76 & 1.11 & 1.33 & 1.83 & 1.93 \\ \hline
\textbf{DI} & 1.40 & 1.34 & 2.26 & 1.85 & 3.64 & 2.87 & 2.30 & 2.01 \\ \hline
\end{tabular}}
\end{table}

\subsection{Experimental results} 
\begin{figure}[!hpt]
\centering
    \includegraphics[width=0.45\textwidth]{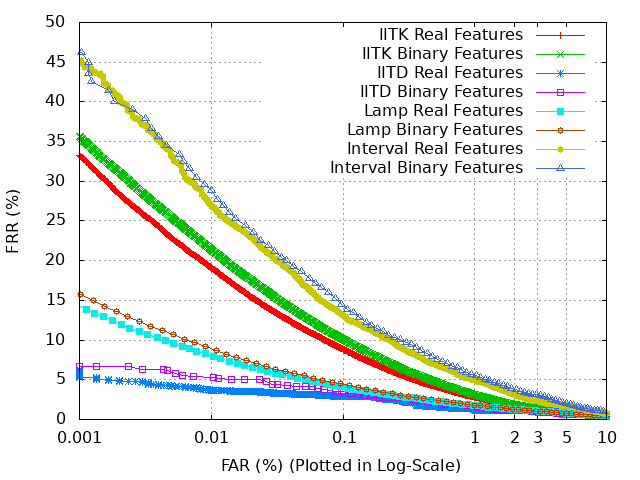}
  \caption{ROC curves of the proposed system on Lamp, Interval, IITD, IITK dataset\label{roc_figure}}
\end{figure}

For experimentally validating the performance of our proposed approach, we have evaluated recognition as well as indexing performance of the proposed technique over four public datasets. Table \ref{tab_real} shows the recognition performance of real-valued feature vectors extracted from POFNet and recognition performance of iris barcodes obtained after binarizing real-valued feature vectors .  It is indicated by Table \ref{tab_real}  that once the real value feature vectors are binarized to form iris barcodes, the recognition performance does not degrade conspicuously (justified discrimination preserving assurance). Figure.\ref{roc_figure} shows the ROC curve on various datasets depicting the trade-off between real-valued iris features and binarized features. It is evident that iris barcodes are close to real value feature vectors (in terms of discrimination) for every dataset.

\textbf{Recognition performance comparative analysis:} To the best of our knowledge~\cite{latest_iris} is the current state-of-the-art iris indexing paper. Here, in this state-of-art ~\cite{latest_iris} paper recognition performance is evaluated with the same training and testing protocol as ours, but for five different types of feature descriptors and EER is evaluated for different types of segmentation errors separately. Thus, for comparative analysis, we have taken their average. Table \ref{tab_per} shows the comparative analysis; it is clearly evident that our method outperforms the current-state-of-art ~\cite{latest_iris} in terms of EER for both the datasets  and comparable results for other performance parameters. \color{black} Although for Interval dataset ~\cite{latest_iris} yields better recall and precision than ours but it should be noted that  recall and precision as individuals are not a good measure for deciding the accuracy of a biometric system. In biometrics, we have number of imposter attempts several times greater than  number of genuine attempts, but the precision term actually measures the probability of correct detection of genuine samples, it is more focused towards genuine class and thus, not affected by  presence of large number of imposter matching. Hence,  EER should be looked upon which depends on both FAR as well as on   FRR.
 \color{black}

\begin{table}[]
\centering
\small{
\caption{Recognition performance comparative analysis}
\label{tab_per}
\begin{tabular}{|l|l|l|l|l|}
\hline
\multirow{2}{*}{\textbf{Method}} & \multicolumn{1}{c|}{\multirow{2}{*}{\textbf{Dataset}}} & \multicolumn{3}{c|}{\textbf{Parameter}} \\ \cline{3-5} 
 & \multicolumn{1}{c|}{} & \textbf{EER\%} & \textbf{Recall} & \textbf{Precision} \\ \hline
\multirow{2}{*}{Ours} & Interval & 2.53 & 0.9381 & 0.9512 \\ \cline{2-5} 
 & IITD & 1.11 & 0.9777 & 0.9787 \\ \hline
\multirow{2}{*}{~\cite{latest_iris}} & Interval & 2.91 & 0.973006 & 0.979804 \\ \cline{2-5} 
 & IITD & 2.81 & 0.971842 & 0.971878 \\ \hline
\end{tabular}}
\end{table}

\begin{figure}
\centering
    \includegraphics[width=0.85\linewidth]{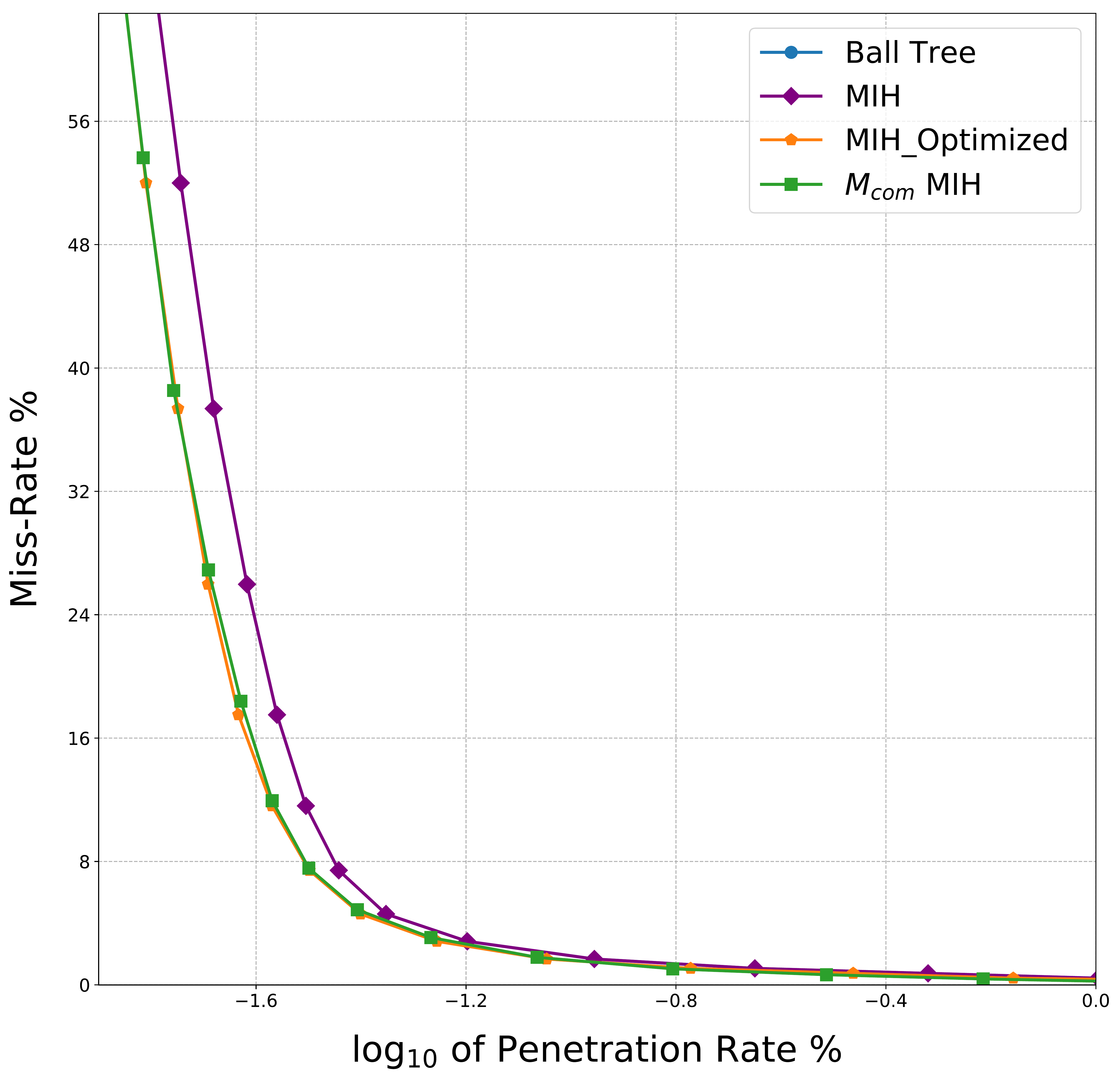}
  \caption{Miss rate vs log$_{10}$ of Penetration Rate \% curves  on IITK dataset\label{hitpen_figure}}
\end{figure}  
\begin{figure}
\centering
    \includegraphics[width=0.85\linewidth]{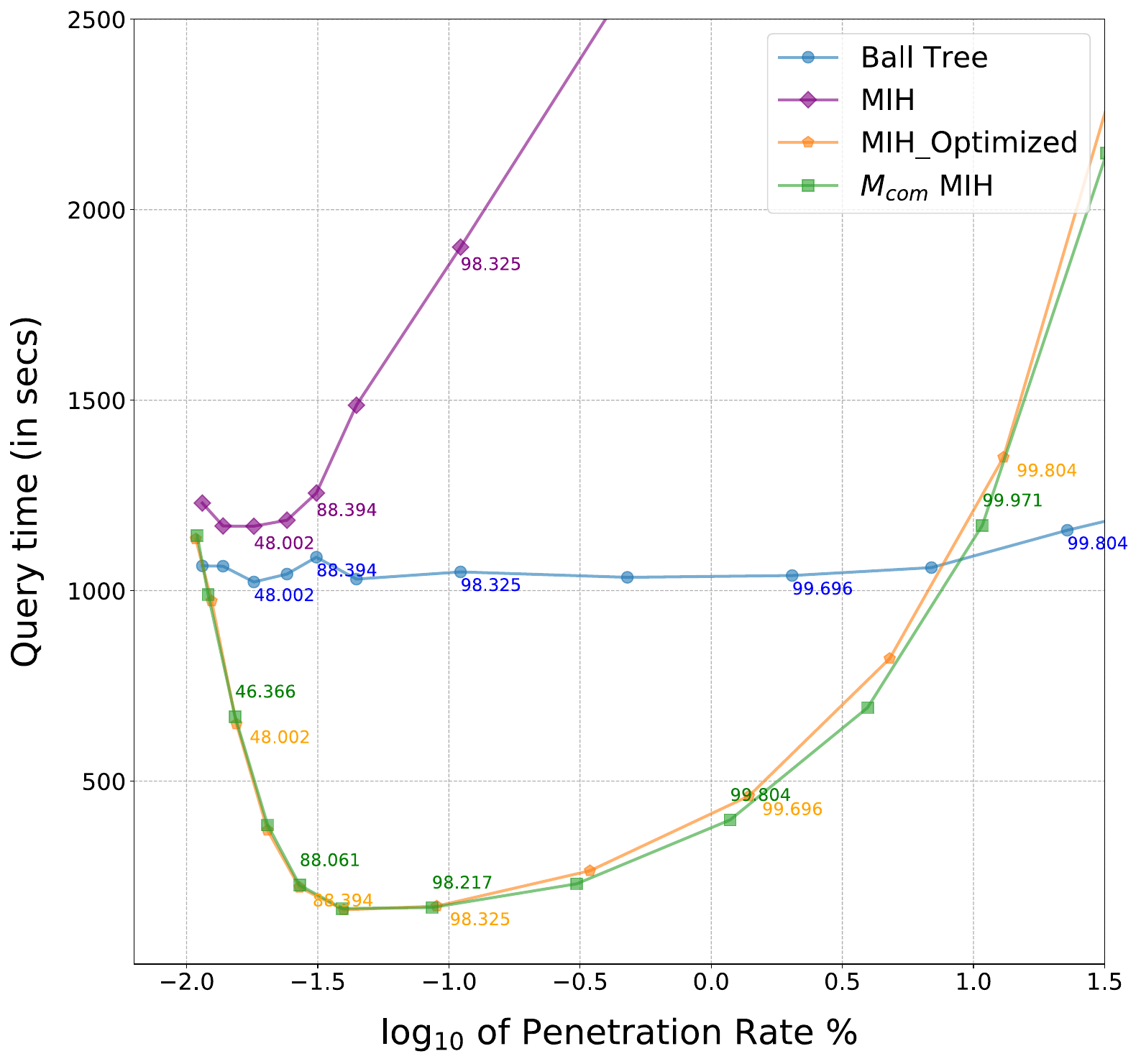}
  \caption{Query Time vs log$_{10}$ of Penetration Rate \% curves  on IITK dataset. The hit-rate percentage has also been reported for comparison. \label{timepen_figure}}
\end{figure}

\begin{table*}[!ht]
    \centering
    \small{
    \caption{Hit rate, penetration rate and query execution time in sec for different datasets (Note here the best performing performance is reported)}
    \label{}
    \begin{tabular}{|*{13}{l|}}
        \hline
        \multirow{2}{*}{} & \multicolumn{3}{l|}{Lamp} & \multicolumn{3}{l|}{Interval} & \multicolumn{3}{l|}{IITD} & \multicolumn{3}{l|}{IITK} \\
        \cline{2-13}
        & HR & PR & Time(s) & HR & PR & Time(s) & HR & PR & Time(s) & HR & PR & Time(s)\\ \hline
        Ball-tree & 99.82 & 0.87 & 600.5 & 98.72 & 3.06 & 1.04 & 98.66 & 1.21 & 1.05 & 99.56 & 1.01 & 1039.6\\ \hline
        MIH & 99.82 & 0.83 & 1594.2 & 99.00 & 3.06 & 27.75 & 98.66 & 1.21 & 7.80 & 99.56 & 0.87 & 3048.5 \\ \hline
        Opt. MIH & 99.82 & 0.63 & 186.7 & 98.73 & 2.40 & 5.12 & 98.66 & 1.11 & 1.53  & 99.56 & 0.69 & 342.56\\ \hline
        $M_{com}$ MIH & 99.82 & 0.62 & 171.7 & 98.73 & 2.11 & 5.00 & 99.11 & 1.18 & 1.36 & 99.56 & 0.64 & 296.61 \\ \hline
    \end{tabular}
    \label{tab_hit}}
\end{table*}

\textbf{Indexing performance analysis:} It is evaluated in terms of hit rate and penetration rate. 
Table \ref{tab_hit} indicates hit rate(HR), penetration rate(PR) and query time taken in seconds for various datasets. It should be noted that for all the datasets, HR is above $98\%$ while PR is below $1.2\%$. Moreover, while all the indexing techniques give almost similar HR, the PR is quite different. Furthermore, PR is quite high in the case of Ball-tree and naive MIH for all the datasets but almost similar for optimized MIH and $M_{com}$ MIH as depicted in Figure.\ref{hitpen_figure}. The real power of indexing can be evaluated on large datasets only. In our case, IITK is the largest dataset, and by considering  $M_{com}$ MIH indexing strategy PR is almost $20\%$ less than the naive MIH and $40\%$ less than the Ball-tree method as shown in Table \ref{tab_hit}.

\textbf{Indexing performance time analysis:} Figure.\ref{timepen_figure} depicts the relation between query time and penetration rate. The supremacy of $M_{com}$ is evident as it outperforms all the other approaches both in terms of query time and hit rate. It can be observed for a specific range of penetration (from $\sim0.1$ to $\sim4$), Optimized MIH, and $M_{com}$ MIH takes significantly less time than Ball Tree and naive MIH with excellent accuracy. All MIH approaches have additional computational costs and work best with large data. 
The additional computational costs can be used to justify the initial decrease in query time for Optimized MIH and $M_{com}$ MIH approaches with an increase in penetration rate.

\textbf{Indexing performance comparative analysis:} Table \ref{tab_comp1} compares our results with some state-of-the-art methods. It should be noted that in all the cases, different datasets along with different training and testing strategy is used. Although current state-of-the-art paper ~\cite{latest_iris} uses the same training and testing strategy, but unlike our approach they have not used the entire testing dataset as query set.


\begin{table}[]
\centering
\small{
\caption{Comparative analysis with existing work}
\label{tab_comp1}
\begin{tabular}{|p{1.6cm}|p{3cm}|p{3cm}|}
\hline
\textbf{Author} & \textbf{Dataset Used} & \textbf{HR \& PR} \\ \hline
Ahmed~\cite{latest_iris} & IITD,Interval, UBIRIS 2 & HR($95\% -99\%$) for PR ($1\%$ - $5\%$) \\ \hline
 Khalaf~\cite{khalaf}& Interval, IITK, BATH & HR \{1) $69.63\%$, 2)$57.74\%$\} PR 
    \{1) $0.98\%$, 2) $0.13\%$ \} \\ \hline
    Drozdowski ~\cite{rathgeb}  & Combined dataset &  HR $98\%$, PR $10\%$\\ \hline
    Ours &  CASIA Lamp, Interval, IITD, IITK & HR above $98\%$ for PR ($1\%$ - $3\%$)\\ \hline
\end{tabular}}
\end{table}

\section{Conclusion} In this paper, we have proposed a deep network based on domain knowledge using dilobe ordinal filtering (POFNet) to extract patch level iris features. To binarize iris features, we have introduced a lambda layer based on SimHash such that discrimination between generated binary bits is preserved. Further, we have introduced a new loss function $M_{com}$ to balance binary bits that make it compatible with the Multi-Index-Hashing technique. To the best of our knowledge, this is the first attempt in designing an end-to-end network for iris indexing. Further, the experimental results depict the efficacy of our proposed approach.

{\small
\bibliographystyle{ieee}
\bibliography{submission_example}
}

\end{document}